\documentclass[11pt]{article}

\usepackage[preprint]{acl}

\usepackage{times}
\usepackage{latexsym}
\usepackage[T1]{fontenc}
\usepackage[utf8]{inputenc}
\usepackage{microtype}
\usepackage{inconsolata}

\usepackage{amsmath,amsfonts,bm}

\def\eqref#1{equation~\ref{#1}}

\def\1{\bm{1}}

\DeclareMathAlphabet{\mathsfit}{\encodingdefault}{\sfdefault}{m}{sl}
\SetMathAlphabet{\mathsfit}{bold}{\encodingdefault}{\sfdefault}{bx}{n}

\usepackage{amssymb}
\usepackage{booktabs}
\usepackage{array}
\usepackage{multirow}
\usepackage{graphicx}
\usepackage{capt-of}
\usepackage{xcolor}
\usepackage{colortbl}

\newcommand{\methodname}{ETRO}

\title{Distill Until Divergence:\\ Error-Truncated Hindsight Distillation for Agentic RL}
\author{
      Yibin Huang\textsuperscript{1},
      Conghui Zhu\textsuperscript{1}\thanks{Corresponding author.}
      \\
      \textsuperscript{1}Faculty of Computing, Harbin Institute of Technology
  }

\begin{document}
\maketitle
\begin{abstract}
Long-horizon interaction requires agents to make multiple decisions in a changing environment. Outcome-based GRPO lacks fine-grained guidance for intermediate decisions. Recent methods distill completed trajectories into reusable skills and use hindsight-conditioned OPD to provide dense token-level credit. However, skill-induced probability shifts do not necessarily represent reward-aligned credit; especially after an erroneous action changes subsequent interaction states, skill-conditioned supervision may become mismatched with the current state. To address this mismatch, we propose \methodname{}, which routes two complementary hindsight skills according to trajectory outcomes and error boundaries: Positive skills capture successful workflows to reinforce successful trajectories and pre-error decisions, while Negative skills capture failure-specific corrections to target decisive errors. The post-error suffix is optimized exclusively by GRPO, retaining outcome-based learning without extending potentially mismatched hindsight supervision. Extensive experiments demonstrate competitive task performance and support effective fine-grained credit allocation.
\end{abstract}

\section{Introduction}
\label{sec:intro}

Large language model agents are increasingly used for tasks that require sustained interaction with external tools and environments, including web navigation~\citep{wei2025webagent}, household tasks~\citep{shridhar2020alfworld}, and search-augmented reasoning~\citep{huang2026bicaa}. These applications extend language models beyond single-response generation toward long-horizon task execution, where each action can alter the observations and choices available in subsequent turns.

Reinforcement learning provides a natural way to learn these strategies from interaction outcomes, but outcome-based GRPO lacks explicit guidance for intermediate decisions~\citep{shao2024deepseekmath}. Recent methods supplement it with trajectory-derived skills, using on-policy distillation (OPD) to provide token-level credit~\citep{yang2026opid,wu2026seed}. Yet denser credit is not necessarily better-aligned credit: after an error changes the interaction state, the same skill may no longer guide subsequent decisions appropriately. Extending its supervision over the suffix can therefore introduce state-mismatched updates~\citep{tan2026atod,qu2026turnsight}.

Our view is that hindsight should inform selective credit allocation. We refer to guidance extending beyond the states and decisions it meaningfully describes as \emph{hindsight applicability mismatch}. A skill that explains a completed trajectory need not be equally applicable to every decision along it. An unrecovered decisive error can separate workflow consolidation, local correction, and outcome-based learning.

We propose \methodname{} to route two functionally distinct hindsight skills. A Positive skill summarizes a successful workflow and guides OPD throughout successful trajectories and over the pre-error prefix of failed trajectories. A Negative skill summarizes failure-specific corrections and guides OPD only at the writer-estimated first unrecovered decisive error. The post-error suffix remains optimized exclusively by GRPO, removing potentially mismatched auxiliary credit without discarding outcome-based learning. By integrating skill-conditioned OPD into the GRPO objective, \methodname{} combines targeted hindsight guidance with outcome-based learning without requiring skills during task execution.

We evaluate \methodname{} on WebShop, ALFWorld, and Search QA using Qwen2.5 actors from 1.5B to 7B parameters. At 7B, \methodname{} achieves 85.8\% WebShop success, 97.0\% ALFWorld success, and 51.8\% Search QA macro-average accuracy. Ablation studies further support the effectiveness of its selective credit allocation.

\begin{samepage}
Our contributions are threefold:
\begin{itemize}
    \item We formulate \emph{hindsight applicability mismatch}: skill-induced probability shifts need not provide reward-aligned credit after an error changes the interaction state.
    \item We introduce \methodname{}, which separates workflow guidance from failure-specific correction and routes their OPD credit by trajectory region, while retaining outcome-based GRPO over the complete rollout.
    \item Extensive experiments across three agentic domains demonstrate \methodname{}'s effectiveness, with ablation studies validating its selective credit allocation.
\end{itemize}
\end{samepage}

\section{Related Work}
\label{sec:related}

\paragraph{\textbf{Reinforcement Learning with Outcome Rewards.}}
Outcome-reward reinforcement learning optimizes policies from completed-trajectory feedback without requiring intermediate annotations, making it suitable for tasks with automatically verifiable outcomes. PPO stabilizes policy-gradient updates through clipping and a learned value function that estimates expected returns~\citep{schulman2017proximal}. ReMax uses a rollout-derived baseline to reduce the need for a separately trained critic~\citep{li2023remax}. RLOO estimates advantages by comparing each sampled response with a leave-one-out baseline from other responses~\citep{ahmadian2024back}. GRPO derives group-relative advantages from verifiable rewards within a sampling group rather than a learned value model~\citep{shao2024deepseekmath}. DAPO improves the stability and efficiency of group-based policy optimization~\citep{yu2026dapo}. Search-R1 extends outcome-driven training to agents that interleave reasoning and search, using answer-correctness rewards~\citep{jin2025search}. WebAgent-R1 trains interactive web agents using task-success feedback across multi-step interactions~\citep{wei2025webagent}. Rather than changing the outcome-reward estimator, ETRO retains GRPO and supplements trajectory-level feedback with auxiliary hindsight credit routed according to the position and role of intermediate decisions.

\paragraph{\textbf{On-Policy Hindsight and Skill Distillation.}}
On-policy hindsight converts post-rollout information into dense supervision on current-policy samples, using completed experience to guide subsequent policy updates. OPSD constructs self-teachers for distillation on the policy's own sampled responses~\citep{zhao2026self}. SDPO conditions self-teachers on feedback to improve supervision of those responses~\citep{hubotter2026reinforcement}. RLSD constrains hindsight likelihood shifts when modulating reward-driven updates~\citep{yang2026self}. Skill-SD summarizes interaction experience as natural-language skills for actor distillation~\citep{wang2026skill}. OPID uses trajectory-derived skills as privileged context for on-policy distillation~\citep{yang2026opid}. SEED distills skills extracted from completed experience into the agent~\citep{wu2026seed}. TRIAL calibrates hindsight signals through trajectory-relative normalization~\citep{zheng2026trajectory}. ATOD regulates distillation using uncertainty~\citep{tan2026atod}. ICSD calibrates supervision by its influence on the optimization objective~\citep{lan2026trust}. TurnSight assesses turn-level supervision through lookahead agreement~\citep{qu2026turnsight}. ETRO instead distinguishes the temporal applicability of workflow and failure-diagnostic skills, routing them before and at a diagnosed error while disabling auxiliary OPD on the subsequent suffix.
ETRO thus addresses skill applicability, not only supervision strength. Truncating auxiliary OPD leaves outcome-based learning active on the suffix.

\begin{figure*}[t]
\centering
\includegraphics[width=0.88\textwidth]{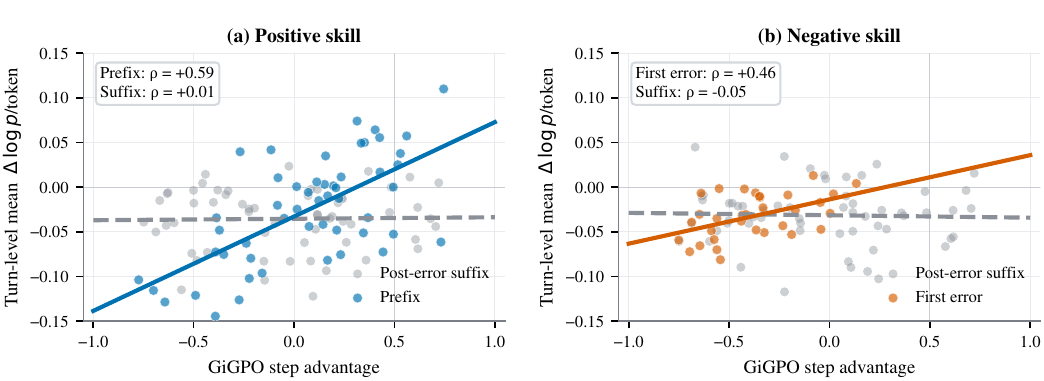}
\caption{OPD signals versus GiGPO-based step advantages on Qwen2.5-1.5B-Instruct failed trajectories. Each point represents one interaction turn.}
\label{fig:preliminary-alignment}
\end{figure*}

\paragraph{\textbf{Localized Supervision and Path Divergence.}}
Localized supervision provides finer-grained evidence about intermediate decisions than terminal rewards alone, distinguishing useful progress from errors within trajectories that share the same final outcome. Math-Shepherd estimates step correctness by sampling continuations from intermediate reasoning states and uses these estimates to train a process reward model~\citep{wang2024math}. OmegaPRM applies divide-and-conquer search to locate the first erroneous reasoning step, turning outcome information into a localized process label~\citep{luo2024improve}. ROSD uses reflection to identify an erroneous span and directs self-distillation toward that span rather than uniformly supervising the entire response~\citep{zhao2026rosd}. HSD conditions on a successful sibling trajectory to concentrate supervision near the point where successful and unsuccessful paths diverge~\citep{li2026localizing}. Beyond localizing supervision, ETRO partitions failed trajectories at a diagnosed unrecovered error into Positive-guided prefixes, Negative-guided error turns, and GRPO-only suffixes.

\section{Method}
\label{sec:method}

\subsection{Overview}

The remainder of this section is organized as follows. Section~\ref{sec:problem} formalizes the multi-turn agent setting, and Section~\ref{sec:motivation} presents the empirical observations that motivate our design. Section~\ref{sec:etro} then introduces the \methodname{} algorithm.

\begin{figure*}[t]
\centering
\includegraphics[width=0.90\textwidth]{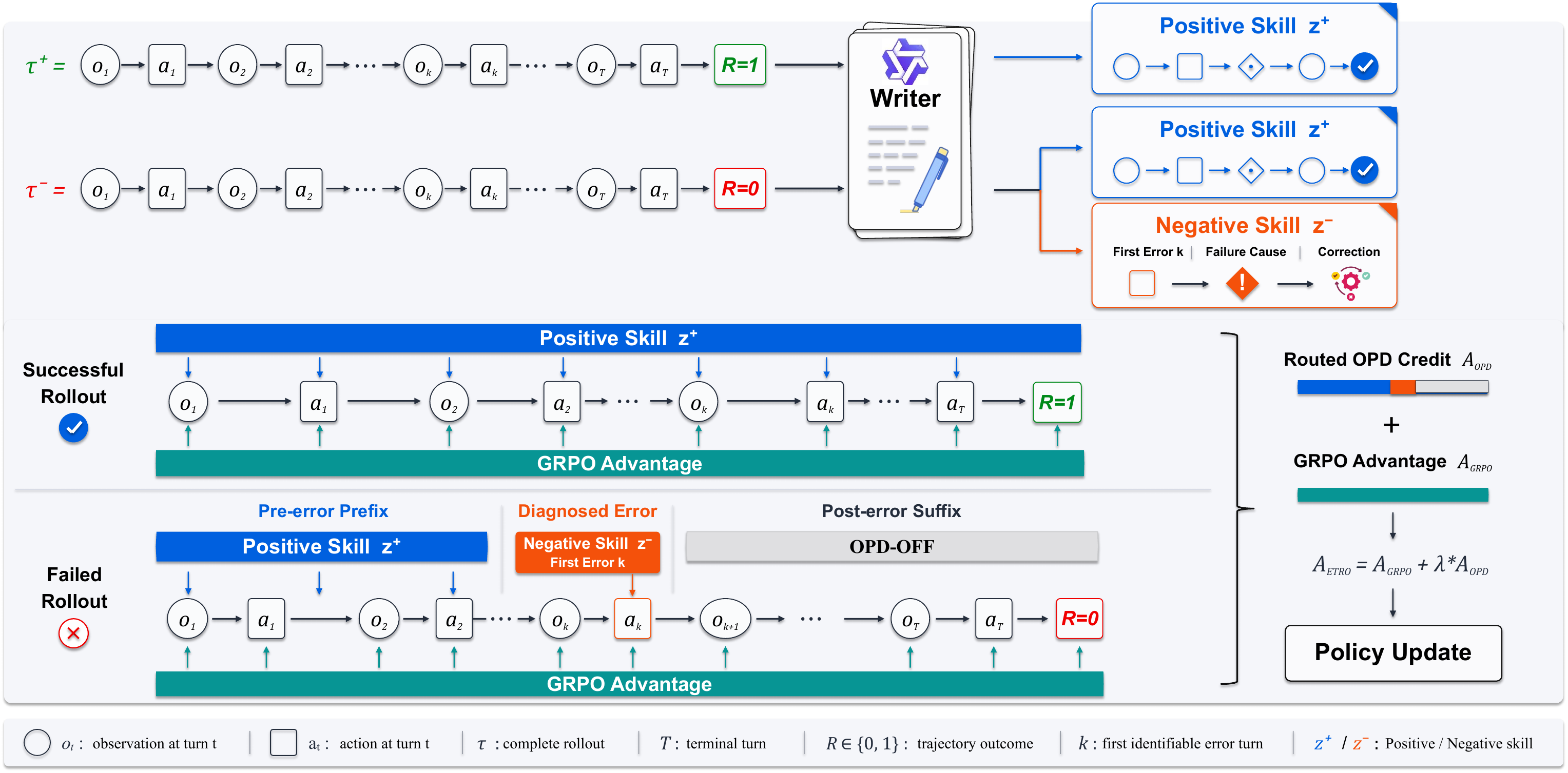}
\caption{Overview of \methodname's trajectory-aware routing of Positive and Negative hindsight credit.}
\label{fig:overview}
\end{figure*}

\subsection{Problem Formulation}
\label{sec:problem}

Let $\mathcal{D}$ be a collection of tasks and $\mathcal{E}$ an interactive environment or external tool. Given a task $x\sim\mathcal{D}$ and an initial observation $o_1$, the agent interacts with $\mathcal{E}$ to produce a trajectory $\tau=(x,o_1,u_1,\ldots,u_T)$, where $T$ is the number of interaction turns. Each turn $u_t=(y_t,a_t,o_{t+1})$ consists of the model-generated response $y_t$, the executable action $a_t$ extracted from that response, and the resulting environment observation $o_{t+1}$. The response contains reasoning and a task-specific action or final answer, generated by the policy $\pi_\theta$ conditioned on the task and preceding interaction history. The trajectory ends with a final answer, environment completion, or exhaustion of the turn budget. A task-specific verifier assigns a binary training outcome indicating success or failure. For trajectory $i$, we use $\tau_i$ and $T_i$ to denote the trajectory and its number of turns, and $y_{i,t,j}$ to denote token $j$ of the model response at turn $t$.

\subsection{Preliminary Study: Regional Alignment of OPD Credit}
\label{sec:motivation}
\label{sec:prelim-alignment}

We investigate whether the alignment between skill-induced OPD signals and outcome-based action quality varies across trajectory regions. A GRPO-trained Qwen2.5-1.5B-Instruct actor samples eight complete trajectories per task prompt. The writer summarizes reusable workflows from successful and failed trajectories as Positive skills and failure-specific corrections as Negative skills. On failed trajectories, we quantify each skill's influence by averaging the difference between skill-conditioned and unconditioned log-probabilities over each turn's reasoning and action tokens. We assess alignment through Spearman correlations with step-level advantages derived from GiGPO's state--action grouping approach~\citep{feng2026group}, separately before, at, and after the diagnosed error. These advantages compare pooled action returns against a shared-state baseline; estimation and filtering details appear in Appendix~\ref{sec:alignment-protocol}.

OPD modulation shows region-dependent correlations with the action-pooled reference advantage, with stronger positive associations in the prefix for Positive skills and at the diagnosed error for Negative skills than in the post-error suffix. As shown in Figure~\ref{fig:preliminary-alignment}, Positive modulation has a correlation of $0.59$ in the prefix but $0.01$ in the suffix; Negative modulation has a correlation of $0.46$ at the diagnosed error but $-0.05$ in the suffix. The weak suffix correlations suggest that, after an error changes the interaction state, skill-induced probability shifts may no longer align with outcome-based action quality. This potential mismatch motivates testing Positive guidance before the error and Negative guidance at the error, while removing suffix OPD to avoid extending potentially mismatched skill supervision and retaining GRPO to learn from the trajectory's actual outcome.

\subsection{Error-Truncated Polarity-Routed Distillation}
\label{sec:etro}

Figure~\ref{fig:overview} illustrates how \methodname{} routes hindsight supervision using a writer-estimated error boundary. For each completed trajectory $\tau_i$, the writer produces a Positive skill $z_i^+$ that summarizes reusable task-solving strategies to reinforce effective decisions through OPD. For a failed trajectory, it also identifies the earliest turn $k_i$ whose erroneous decision is judged decisive for the observed failure and remains uncorrected by later actions, and produces a Negative skill $z_i^-$ describing the decision, its consequence, and an executable alternative. This boundary separates Positive-guided decisions before the error from the local Negative correction at the error and the GRPO-only suffix. Failed trajectories without a valid boundary retain Positive guidance throughout; we represent this case by $k_i=\infty$. Writer prompts, boundary validation, and skill construction details appear in Appendix~\ref{sec:writer-schema}.

Let $c_{i,t,j}$ denote the original generation context of token $y_{i,t,j}$, comprising the task, preceding interaction history, and earlier tokens in the current response. Let $p\in\{+,-\}$ index the skill polarity, with $z_i^p$ denoting the corresponding skill. We compute the OPD gap:
\begin{equation}
\begin{aligned}
\Delta^p_{i,t,j}
&=\log \pi_{\mathrm{old}}(y_{i,t,j}\mid c_{i,t,j},z_i^p)\\
&\quad-\log \pi_{\mathrm{old}}(y_{i,t,j}\mid c_{i,t,j}),
\end{aligned}
\label{eq:opd-gap}
\end{equation}
Here, $\pi_{\mathrm{old}}$ is the frozen rollout policy, and $\Delta^p_{i,t,j}$ is fixed during optimization. Skills provide additional context only for re-scoring and are absent during rollout and evaluation.

The unweighted routed advantage for each eligible reasoning or action token is
\begin{equation}
A^{\mathrm{OPD}}_{i,t,j}=
\begin{cases}
\Delta^+_{i,t,j}, & \text{if $\tau_i$ succeeds},\\
\Delta^+_{i,t,j}, & \text{if $\tau_i$ fails and $t<k_i$},\\
\Delta^-_{i,t,j}, & \text{if $\tau_i$ fails and $t=k_i$},\\
0, & \text{otherwise}.
\end{cases}
\label{eq:routed-credit}
\end{equation}
We weight this auxiliary advantage by $\lambda$ and add it to the trajectory-level GRPO advantage:
\begin{equation}
A^{\mathrm{ETRO}}_{i,t,j}
=A_i^{\mathrm{GRPO}}+\lambda A^{\mathrm{OPD}}_{i,t,j}.
\label{eq:totaladv}
\end{equation}
Here, $A_i^{\mathrm{GRPO}}$ is the group-relative outcome advantage of trajectory $i$, and $\lambda$ controls the strength of both OPD branches. We substitute $A^{\mathrm{ETRO}}_{i,t,j}$ into the standard clipped GRPO objective, setting the OPD term to zero outside eligible reasoning and action spans. Post-error truncation therefore removes only auxiliary supervision; GRPO continues to optimize every valid response token in the complete rollout.

\section{Experiments}
\label{sec:experiments}

\begin{table*}[t]
\caption{Results on ALFWorld, Search QA, and WebShop across Qwen2.5-Instruct model scales (\%). Higher values indicate better performance.}
\label{tab:main}
\centering
\footnotesize
\renewcommand{\arraystretch}{1.12}
\setlength{\tabcolsep}{3pt}
\begin{tabular*}{\textwidth}{@{\extracolsep{\fill}}>{\centering\arraybackslash}m{2.1cm}*{11}{c}@{}}
\toprule
\multirow{2}{*}{\textbf{Method}} & \multicolumn{1}{c}{\textbf{ALFWorld}} & \multicolumn{8}{c}{\textbf{Search QA}} & \multicolumn{2}{c}{\textbf{WebShop}} \\
\cmidrule(lr){2-2}\cmidrule(lr){3-10}\cmidrule(lr){11-12}
& \textbf{Succ.} & \textbf{NQ} & \textbf{Triv} & \textbf{Pop} & \textbf{Hotp} & \textbf{2Wk} & \textbf{MuS} & \textbf{Bam} & \textbf{Avg.} & \textbf{Score} & \textbf{Succ.} \\
\midrule
\rowcolor{gray!15}
\multicolumn{12}{@{}l}{\textbf{Qwen2.5-1.5B-Instruct}} \\
\midrule
GRPO & 65.7 & 16.6 & 34.1 & 15.9 & 16.1 & 24.8 & 3.4 & 58.1 & 24.1 & 72.0 & 62.0 \\
GRPO+OPSD & 71.1 & 16.9 & 34.3 & 16.4 & 16.3 & 24.8 & 3.4 & 56.9 & 24.1 & 84.9 & 73.4 \\
Skill-SD & 68.0 & 16.7 & 35.4 & 14.8 & 16.7 & 25.2 & 3.9 & 11.2 & 17.7 & 80.6 & 68.8 \\
RLSD & 67.2 & 13.9 & 29.4 & 12.4 & 14.2 & 21.1 & 3.0 & 9.5 & 14.8 & 80.5 & 63.3 \\
SDAR & 72.7 & 16.1 & 32.7 & 15.8 & 15.9 & 24.4 & 3.3 & 4.4 & 16.1 & 81.9 & 64.1 \\
ICSD & 73.4 & 18.3 & 39.2 & 17.1 & 18.7 & 28.2 & 4.4 & 68.0 & 27.7 & 85.6 & 75.8 \\
\textbf{\methodname{}} & \textbf{79.1} & \textbf{19.4} & \textbf{40.8} & \textbf{18.7} & \textbf{20.1} & \textbf{30.4} & \textbf{5.2} & \textbf{69.8} & \textbf{29.2} & \textbf{87.9} & \textbf{80.8} \\
\midrule
\rowcolor{gray!15}
\multicolumn{12}{@{}l}{\textbf{Qwen2.5-3B-Instruct}} \\
\midrule
GRPO & 75.0 & 39.3 & 60.6 & 41.1 & 37.4 & 34.6 & 15.4 & 26.4 & 36.4 & 79.8 & 64.4 \\
GRPO+OPSD & 81.2 & 44.9 & 61.2 & 45.2 & 40.4 & 38.5 & 16.0 & 66.1 & 44.6 & 77.8 & 66.4 \\
Skill-SD & 73.4 & 44.4 & 60.4 & 44.0 & 39.5 & 40.4 & 15.4 & 64.9 & 44.1 & 75.9 & 64.0 \\
RLSD & 79.7 & 41.5 & 58.6 & 42.3 & 40.4 & 40.2 & 16.8 & 66.9 & 43.8 & 84.4 & 66.4 \\
SDAR & 84.4 & 44.8 & 58.1 & 44.3 & 38.6 & 36.2 & 15.7 & 66.1 & 43.4 & 85.0 & 68.0 \\
OPID & 84.3 & 45.9 & 61.4 & 45.7 & 40.7 & 38.8 & 16.4 & 66.1 & 45.0 & 85.0 & 74.2 \\
SEED & 91.8 & 44.3 & 61.9 & 47.2 & 41.0 & 42.0 & 16.9 & 67.7 & 45.9 & 88.5 & 78.9 \\
\textbf{\methodname{}} & \textbf{93.3} & \textbf{47.2} & \textbf{63.8} & \textbf{48.5} & \textbf{43.2} & \textbf{44.1} & \textbf{18.7} & \textbf{70.8} & \textbf{48.0} & \textbf{90.3} & \textbf{83.0} \\
\midrule
\rowcolor{gray!15}
\multicolumn{12}{@{}l}{\textbf{Qwen2.5-7B-Instruct}} \\
\midrule
GRPO & 81.2 & 45.1 & 63.7 & 44.0 & 43.6 & 43.2 & 16.8 & 37.6 & 42.0 & 80.9 & 73.7 \\
GRPO+OPSD & 80.4 & 47.3 & 64.5 & 46.9 & 43.8 & 39.3 & 18.0 & 69.4 & 47.0 & 86.8 & 76.5 \\
Skill-SD & 85.1 & 47.1 & 64.5 & 47.8 & 44.2 & 42.1 & 20.2 & 69.0 & 47.8 & 86.1 & 76.5 \\
RLSD & 82.0 & 46.8 & 63.0 & 44.4 & 45.5 & 48.9 & 21.5 & 73.0 & 49.0 & 87.4 & 77.3 \\
SDAR & 85.9 & 46.3 & 63.5 & 48.2 & 43.8 & 48.4 & 19.6 & 73.0 & 49.0 & 89.4 & 82.8 \\
OPID & 90.0 & 48.8 & 65.6 & 46.8 & 46.1 & 42.7 & 21.7 & 72.6 & 49.2 & 85.3 & 79.7 \\
SEED & 96.1 & 47.0 & 64.9 & 47.8 & 45.2 & 45.3 & 20.1 & 70.2 & 48.6 & 89.7 & 78.1 \\
\textbf{\methodname{}} & \textbf{97.0} & \textbf{50.4} & \textbf{67.1} & \textbf{49.5} & \textbf{48.0} & \textbf{50.2} & \textbf{23.1} & \textbf{74.0} & \textbf{51.8} & \textbf{91.7} & \textbf{85.8} \\
\bottomrule
\end{tabular*}
\end{table*}

\subsection{Experimental Setup}
\label{sec:setup}

\paragraph{Benchmarks and metrics.}
We evaluate \methodname{} on WebShop~\citep{yao2022webshop}, ALFWorld~\citep{shridhar2020alfworld}, and Search QA. We report normalized score and exact success for WebShop, success rate for ALFWorld, and average answer accuracy across the seven Search QA datasets.

\paragraph{Models and baselines.}
We train Qwen2.5-Instruct models at 1.5B, 3B, and 7B scales and compare against GRPO, GRPO+OPSD, Skill-SD, RLSD, SDAR, OPID~\citep{yang2026opid}, SEED~\citep{wu2026seed}, and ICSD~\citep{lan2026trust}. In Table~\ref{tab:main}, ETRO results come from our experiments, whereas all baseline results are taken from prior reports rather than locally reproduced under a unified protocol. Skills are absent during ETRO evaluation. Result provenance and our training and evaluation settings are described in Appendix~\ref{sec:implementation-details}.

\paragraph{Training configuration.}
We sample eight rollouts per task and optimize with AdamW at learning rate $10^{-6}$, using one PPO epoch per batch and clipping parameter 0.2. A frozen Qwen3.5-27B writer supplies skills, with OPD coefficient $\lambda=0.001$.

\paragraph{Evaluation protocol.}
WebShop evaluation covers 500 tasks and ALFWorld uses 134 Unseen tasks, both with a 15-turn limit. Search QA follows the Search-R1 splits and reports normalized exact-match accuracy averaged equally across seven datasets.

\begin{table*}[t]
\begin{minipage}[t]{0.485\textwidth}
\captionof{table}{Skill selection and branch combination on WebShop and ALFWorld using Qwen2.5-1.5B-Instruct.}
\label{tab:polarity-ablation}
\centering
\footnotesize
\renewcommand{\arraystretch}{1.10}
\setlength{\tabcolsep}{5pt}
\begin{tabular}{@{}cccc@{}}
\toprule
\multicolumn{1}{c}{\multirow{2}{*}{\textbf{Method}}} & \multicolumn{2}{c}{\textbf{WebShop}} & \textbf{ALFWorld} \\
\cmidrule(lr){2-3}\cmidrule(lr){4-4}
& \textbf{Succ.} & \textbf{Score} & \textbf{Succ.} \\
\midrule
GRPO & 62.0 & 72.0 & 65.7 \\
Global Positive & 77.2 & 85.3 & 76.9 \\
Global Negative & 68.8 & 75.3 & 72.4 \\
Global Both & 76.0 & 83.8 & 76.9 \\
Positive Only & 77.6 & 85.0 & 77.6 \\
Negative Only & 73.4 & 79.8 & 74.6 \\
\methodname{} & 80.8 & 87.9 & 79.1 \\
\bottomrule
\end{tabular}
\end{minipage}
\hfill
\begin{minipage}[t]{0.485\textwidth}
\captionof{table}{Suffix supervision and boundary selection on WebShop and ALFWorld using Qwen2.5-1.5B-Instruct.}
\label{tab:truncation-ablation}
\centering
\footnotesize
\renewcommand{\arraystretch}{1.10}
\setlength{\tabcolsep}{5pt}
\begin{tabular}{@{}cccc@{}}
\toprule
\multicolumn{1}{c}{\multirow{2}{*}{\textbf{Method}}} & \multicolumn{2}{c}{\textbf{WebShop}} & \textbf{ALFWorld} \\
\cmidrule(lr){2-3}\cmidrule(lr){4-4}
& \textbf{Succ.} & \textbf{Score} & \textbf{Succ.} \\
\midrule
GRPO & 62.0 & 72.0 & 65.7 \\
Suffix Positive & 77.6 & 84.6 & 77.6 \\
Suffix Negative & 72.2 & 79.3 & 74.6 \\
Random Boundary & 72.8 & 79.5 & 73.1 \\
\methodname{} & 80.8 & 87.9 & 79.1 \\
\bottomrule
\end{tabular}
\end{minipage}
\end{table*}

\subsection{Main Results}
\label{sec:results}

Compared with GRPO and its self-distillation extension GRPO+OPSD, ETRO improves performance across all three domains and model scales in Table~\ref{tab:main}. At 1.5B, its WebShop success reaches 80.8\%, exceeding GRPO by 18.8 percentage points. At 3B, its Search QA macro-average accuracy reaches 48.0\%, surpassing GRPO+OPSD by 3.4 points.

Compared with related distillation methods, ETRO also achieves the highest reported results among the listed methods at each scale. At 1.5B, its WebShop and ALFWorld success rates exceed ICSD by 5.0 and 5.7 points; at 3B, they exceed SEED by 4.1 and 1.5 points, respectively. At 7B, ETRO improves over SDAR's WebShop success by 3.0 points and OPID's Search QA macro-average accuracy by 2.6 points. These gains extend beyond outcome-only training to existing distillation approaches across interactive task execution and search-augmented reasoning.

\subsection{Ablation Studies}
\label{sec:ablations}

\noindent\textbf{Skill Selection and Branch Combination.} To assess skill selection and branch combination, we organize the Qwen2.5-1.5B-Instruct comparisons on WebShop and ALFWorld into five groups (Table~\ref{tab:polarity-ablation}): GRPO without OPD; Global Positive and Global Negative with one skill applied throughout the trajectory; Global Both with both skills combined globally; Positive Only and Negative Only with one branch removed from ETRO's routing rule; and the complete ETRO method. Positive Only retains ETRO's Positive guidance on successful trajectories, failed-trajectory prefixes, and complete failed trajectories without a valid boundary, while Negative Only retains its Negative guidance at diagnosed errors. All OPD variants use the same coefficient $\lambda=0.001$ and retain GRPO over the complete rollout; branch removal only changes the coverage of auxiliary supervision.

The complete method outperforms all variants on the three reported metrics, reaching 80.8\% WebShop success and 79.1\% ALFWorld success. On WebShop, it exceeds Global Positive and Global Negative by 3.6 and 12.0 percentage points. Removing the Negative or Positive branch lowers WebShop success to 77.6\% or 73.4\% and ALFWorld success to 77.6\% or 74.6\%, respectively, supporting the complementary roles of workflow guidance and local correction. Global Both averages the two skill-induced gaps throughout failed trajectories while retaining Positive guidance on successful trajectories. Its WebShop success of 76.0\% remains below ETRO, suggesting that combining skills globally is less effective than routing them by trajectory region.

\noindent\textbf{Suffix Supervision and Boundary Selection.} To assess where supervision should end once both branches are available, we compare ETRO with Suffix Positive, Suffix Negative, and Random Boundary using Qwen2.5-1.5B-Instruct (Table~\ref{tab:truncation-ablation}). The two suffix variants keep ETRO's Positive prefix and Negative error-turn guidance but continue the corresponding skill after the diagnosed error. Random Boundary keeps the same Positive--Negative--Off routing structure but partitions each eligible failed trajectory at a randomly selected turn instead of the diagnosed error. All variants retain GRPO over the complete rollout and the same OPD coefficient.

Truncation at the diagnosed error outperforms continued suffix supervision and random-boundary routing on all reported metrics (Table~\ref{tab:truncation-ablation}). On WebShop, ETRO exceeds Suffix Positive and Suffix Negative by 3.2 and 8.6 percentage points in success. Random Boundary reaches 72.8\% success, falling below ETRO by 8.0 points despite also removing suffix OPD. The consistent improvements in WebShop score and ALFWorld success further favor truncation at the diagnosed boundary over extending supervision or selecting an arbitrary cutoff. Section~\ref{sec:analysis} separately examines the placement of local Negative correction.

\begin{figure*}[t]
\centering
\includegraphics[width=0.98\textwidth]{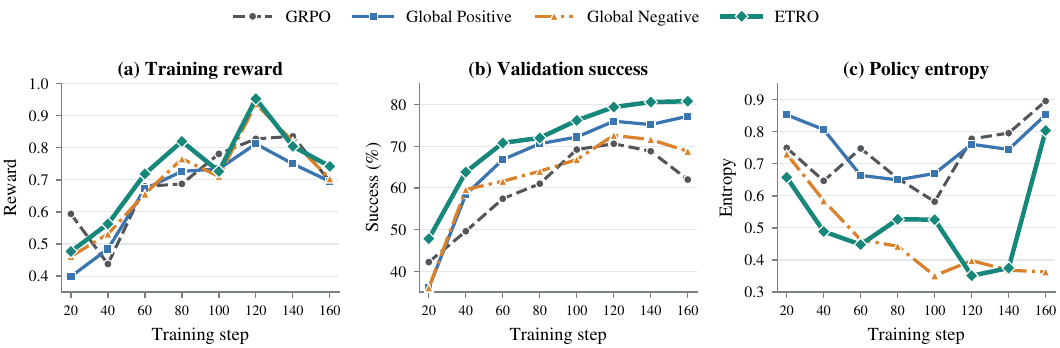}
\caption{Qwen2.5-1.5B-Instruct training dynamics on WebShop for pure GRPO, full-trajectory Positive and Negative OPD, and \methodname{}. Training reward is reported in $[0,1]$. Validation uses a 15-turn interaction limit.}
\label{fig:training-dynamics}
\end{figure*}

\noindent\textbf{Strength of Suffix Supervision.} To distinguish a benefit from weakening suffix supervision from a benefit of removing it, we retain diagnosis-based boundary selection and vary the post-error OPD coefficient as $\alpha\lambda$, with $\alpha\in\{0,0.25,0.5,1\}$ and $\lambda=0.001$, using Qwen2.5-1.5B-Instruct on WebShop (Table~\ref{tab:suffix-opd-weight}). We evaluate both Positive and Negative suffix supervision. Prefix and error-turn routing, successful-trajectory supervision, and the no-boundary fallback remain unchanged. All arms use the same initialization, training settings, writer, boundary-selection procedure, evaluation protocol, and loss normalization. The shared $\alpha=0$ endpoint is ETRO; $\alpha=1$ corresponds to full-strength Suffix Positive or Suffix Negative.

Complete truncation achieves the highest success rate and score in both sweeps (Table~\ref{tab:suffix-opd-weight}). Reducing $\alpha$ from 1 to 0.25 raises WebShop success from 77.6\% to 79.0\% for Positive suffix guidance and from 72.2\% to 76.6\% for Negative guidance. Both trail ETRO's 80.8\% by 1.8 and 4.2 points. Both variants also remain below ETRO in task score. Thus, weakening suffix supervision improves both variants relative to full-strength supervision, but none of the tested nonzero weights matches complete removal.

\begin{table}[t]
\caption{Post-error OPD weight sweep on WebShop using Qwen2.5-1.5B-Instruct, with $\lambda=0.001$.}
\label{tab:suffix-opd-weight}
\centering
\footnotesize
\renewcommand{\arraystretch}{1.12}
\setlength{\tabcolsep}{6pt}
\begin{tabular}{@{}cccc@{}}
\toprule
\multicolumn{1}{c}{\textbf{Method}} & $\alpha$ & \textbf{Success (\%)} & \textbf{Score} \\
\midrule
\methodname{} & 0 & 80.8 & 87.9 \\
\midrule
Suffix Positive & 0.25 & 79.0 & 85.1 \\
Suffix Positive & 0.5 & 78.4 & 85.2 \\
Suffix Positive & 1 & 77.6 & 84.6 \\
\midrule
Suffix Negative & 0.25 & 76.6 & 82.7 \\
Suffix Negative & 0.5 & 74.4 & 80.6 \\
Suffix Negative & 1 & 72.2 & 79.3 \\
\bottomrule
\end{tabular}
\end{table}

\subsection{Analysis Experiments}
\label{sec:analysis}

\noindent\textbf{Training behavior.} To examine whether ETRO's final gains are accompanied by sustained improvement during optimization, we compare GRPO, Global Positive, Global Negative, and ETRO using Qwen2.5-1.5B-Instruct on WebShop. We track unsmoothed training reward, validation success, and policy entropy at the same recorded checkpoints (Figure~\ref{fig:training-dynamics}), assessing task performance alongside changes in policy concentration.

ETRO maintains the highest validation success at every recorded checkpoint without a monotonic decrease in policy entropy. Its success rises from 47.8\% at step 20 to 80.8\% at step 160, exceeding the final Global Positive, Global Negative, and GRPO results by 3.6, 12.0, and 18.8 percentage points, respectively. Training reward reaches 0.9531 at step 120 and fluctuates afterward, while validation success continues to improve. Over the same late-stage interval, ETRO's entropy recovers from 0.351 to 0.804, whereas Global Negative finishes at 0.363 with lower success. These observations show that the reported performance gains coexist with sustained validation improvement and a recovery in response diversity, rather than uniformly increasing policy concentration.

\noindent\textbf{Negative-skill placement.} To examine whether failure-specific guidance benefits from placement at the decision it describes, we vary the location of the one-turn Negative OPD mask using Qwen2.5-1.5B-Instruct on WebShop. For failed trajectories with nonempty prefixes and suffixes, the writer produces a Negative skill for the validated error at turn $k$. Holding that skill's content fixed, we apply the mask either at $k$, at a randomly sampled turn from the pre-error prefix, or at a randomly sampled turn from the post-error suffix. Only the location of Negative supervision varies; the supervised turn count remains fixed.

\begin{figure}[t]
\centering
\includegraphics[width=\linewidth]{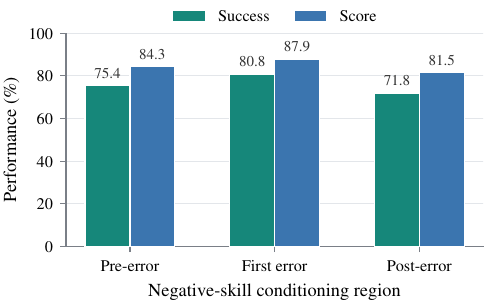}
\caption{Comparison of Negative-skill conditioning regions on WebShop using Qwen2.5-1.5B-Instruct.}
\label{fig:negative-position}
\end{figure}

Applying Negative guidance at the diagnosed error achieves the highest success and score among the three tested placements (Figure~\ref{fig:negative-position}). Error-turn conditioning reaches 80.8\% success and an 87.9 score, compared with 75.4\% and 84.3 for prefix placement and 71.8\% and 81.5 for suffix placement. The corresponding success gains are 5.4 and 9.0 percentage points, with score gains of 3.6 and 6.4. These results favor localizing the correction signal at the diagnosed decision rather than shifting it to an earlier or later turn.

\noindent\textbf{Writer reliability.} To assess the quality of the skills and error boundaries that support routing, we evaluate 100 failed WebShop trajectories sampled after Qwen2.5-1.5B-Instruct training. Five annotators each inspect 20 distinct complete trajectories to establish reference annotations. Using these human annotations as reference answers, GPT-5.6 sol evaluates the writer outputs for boundary acceptance, exact-turn agreement, workflow validity, and correction validity. Results appear in Table~\ref{tab:writer-audit}.

The reference-based evaluation supports the quality of the writer-generated skills and error boundaries on the sampled WebShop trajectories. Boundary acceptance reaches 92.0\%, with 84.0\% exact-turn agreement against human references; workflow and correction validity reach 92.0\% and 90.0\%, respectively. These results support the writer's ability to identify relevant error boundaries and produce actionable guidance for both OPD branches, providing empirical support for the skill generation and localization underlying ETRO's routing.

\begin{table}[t]
\caption{Writer quality on WebShop, evaluated against human references by GPT-5.6 sol (\%).}
\label{tab:writer-audit}
\centering
\footnotesize
\renewcommand{\arraystretch}{1.10}
\setlength{\tabcolsep}{8pt}
\begin{tabular}{@{}cc@{}}
\toprule
\textbf{Metric} & \textbf{Rate (\%)} \\
\midrule
Boundary acceptance & 92.0 \\
Exact-turn agreement & 84.0 \\
Workflow validity & 92.0 \\
Correction validity & 90.0 \\
\bottomrule
\end{tabular}
\end{table}

\noindent\textbf{Post-error recovery.} To assess whether suffix truncation improves behavior after a recoverable error, we compare actors from the same perturbed states on held-out WebShop tasks, collecting repeated continuations without hindsight skills. The actors retain Positive prefix and Negative error-turn guidance during training but differ in suffix treatment: Suffix Positive and Suffix Negative continue the corresponding OPD signal, while ETRO retains only GRPO. We report recovery success, normalized score, wrong-branch continuation, and first-turn action entropy; the controlled-error construction and evaluation procedure appear in Appendix~\ref{sec:recovery-protocol}.

ETRO achieves the highest recovery success and score and the lowest wrong-branch continuation rate among the compared actors (Table~\ref{tab:actor-behavior}). It reaches 63.8\% success and an 82.6 score, compared with 54.6\% and 74.8 for Suffix Positive and 49.2\% and 69.7 for Suffix Negative. Wrong-branch continuation falls to 23.7\%, versus 33.9\% and 40.6\%, reductions of 10.2 and 16.9 percentage points. ETRO's first-turn entropy is 0.62, compared with 0.55 and 0.39 for the suffix variants. These results indicate that removing post-error skill supervision while retaining GRPO helps agents recover from errors rather than persist along an incorrect action branch.
Without hindsight skills at evaluation, these differences reflect learned behavior. Higher entropy accompanies better recovery and fewer wrong-branch continuations, rather than action variability alone.

\begin{table}[t]
\caption{Post-error recovery on WebShop.}
\label{tab:actor-behavior}
\vspace{0.35em}
\centering
\footnotesize
\renewcommand{\arraystretch}{1.10}
\setlength{\tabcolsep}{2.2pt}
\begin{tabular*}{\linewidth}{@{\extracolsep{\fill}}ccccc@{}}
\toprule
\multirow{2}{*}{\textbf{Method}} & \textbf{Recovery} & \textbf{Recovery} & \textbf{Wrong} & \multirow{2}{*}{\textbf{Entropy}} \\
& \textbf{Success} & \textbf{Score} & \textbf{Branch} & \\
\midrule
Suffix Positive & 54.6 & 74.8 & 33.9 & 0.55 \\
Suffix Negative & 49.2 & 69.7 & 40.6 & 0.39 \\
\methodname{} & 63.8 & 82.6 & 23.7 & 0.62 \\
\bottomrule
\end{tabular*}
\end{table}

\section{Conclusion}

We presented \methodname{} to address the mismatch between hindsight supervision and evolving interaction states. By separating workflow guidance from local error correction, \methodname{} routes OPD by skill polarity and trajectory position while preserving GRPO throughout the rollout. Experiments on WebShop, ALFWorld, and Search QA demonstrate competitive performance across model scales. Ablations support complementary skill routing and post-error truncation, while recovery experiments show improved recovery and less wrong-branch continuation. These findings suggest that effective hindsight credit allocation depends on both skill content and its temporal applicability. Future work could extend routing to multiple error and recovery boundaries.

\clearpage
\section*{Limitations}

\methodname{} depends on the writer to locate errors and generate hindsight skills. Our audit and ablation studies support the effectiveness of this design, but its performance may vary with the quality of the writer's outputs.

Finally, hindsight generation and policy re-scoring add training-time computation even though skills are absent at inference time. The approach depends on access to complete trajectories and outcome feedback, which may be costly or unavailable in some applications.

\section*{Ethical Considerations}

Training language-model agents to act more effectively can also increase the consequences of unsafe actions. Success in benchmark environments does not establish safety for deployment with real purchases, private information, or unrestricted external tools. Such deployment would require additional safeguards, authorization checks, and task-specific evaluation beyond the scope of this work.

Interaction trajectories and hindsight skills can contain user requests, retrieved text, or other sensitive information. Any release or reuse of these artifacts should respect applicable dataset and model licenses, protect personal information, and preserve the anonymity of review materials. Writer-generated diagnoses are fallible and should not be treated as authoritative judgments of users or annotators.

\section*{AI Assistance Disclosure}

An AI assistant helped edit manuscript wording and formatting, including initial drafts of the Limitations and Ethical Considerations sections. The authors are responsible for verifying the final text, references, and reported results.

\bibliography{references}
\clearpage
\appendix
\section{Writer Prompt and Output Format}
\label{sec:writer-schema}

The frozen writer receives the task, complete trajectory, episode outcome, and candidate step indices. Its JSON output contains \texttt{episode\_summary}, \texttt{episode\_skill}, and \texttt{step\_skills}. The global \texttt{episode\_skill} supplies Positive guidance, while \texttt{step\_skills} provides a local Negative correction for the first unrecovered decisive error in a failed episode, with at most one entry. The prompt uses zero-based step indices; the corresponding one-based turn index in the method is obtained by adding one.

\paragraph{Boundary selection and recovery.}
The writer evaluates an error using the complete later trajectory, rather than selecting the first locally invalid action. An error is recovered when subsequent actions correct its consequences, as supported by later observations and task progress; such an error is excluded from boundary selection even if the episode ultimately fails for another reason. Among multiple errors, the writer selects the earliest candidate that remains unrecovered and is judged decisive for the observed failure. A format violation alone is insufficient: it must pass the same recovery and decisiveness tests. If no such error can be identified confidently, the writer returns an empty \texttt{step\_skills} object and the failed trajectory retains Positive OPD throughout alongside GRPO. This is a diagnosis of the realized episode, not a claim that the selected action makes failure inevitable.

\paragraph{Candidate indices.}
The candidate list contains all interaction steps in the complete trajectory: for a trajectory with $T$ turns, the zero-based list is $\{0,\ldots,T-1\}$. No recorded interaction steps are filtered out before boundary selection, including steps with format-invalid responses. The writer selects only from this complete list, so ``first'' means earliest in the complete trajectory satisfying the unrecovered-decisive-error definition, not earliest within a filtered subset.

\begin{figure*}[p]
\centering
\begingroup
\setlength{\fboxsep}{9pt}
\setlength{\fboxrule}{0.4pt}
\fcolorbox{gray!45}{gray!10}{%
\begin{minipage}{\dimexpr\textwidth-2\fboxsep-2\fboxrule\relax}
\fontsize{9}{11}\selectfont
\ttfamily\raggedright
\setlength{\parskip}{4pt}
\setlength{\parindent}{0pt}
\detokenize{Analyze the following agent episode and return ONLY valid JSON.}\par

\detokenize{This experiment routes hindsight according to its temporal scope:}\par
\detokenize{- episode_skill is global positive guidance and must always describe what the policy should do to complete the task, regardless of whether the sampled episode succeeded or failed.}\par
\detokenize{- step_skills is local negative correction. It is used only for the first unrecovered decisive error in a failed episode and must include a concrete better decision, not merely a prohibition.}\par

\detokenize{{{ROUTING_INSTRUCTION}}}\par

\detokenize{How to interpret action validity:}\par
\detokenize{- `Action valid` is a deterministic FORMAT signal produced by the WebShop action parser, not a judgment made by this analyzer.}\par
\detokenize{- `Action valid: false` means the response violated the required response format: it was missing a complete <think>...</think> block, missing a complete <action>...</action> block, or contained Chinese characters. Treat this as direct evidence of a format error at that step.}\par
\detokenize{- `Action valid: true` means only that those format checks passed. It does NOT prove that the extracted action was admissible, executable, relevant, or semantically correct. Determine those properties from the current observation, admissible actions, and the following environment observation.}\par
\detokenize{- A format error is not automatically the decisive cause of episode failure. Apply the same recovery and decisiveness tests above using the complete later trajectory.}\par

\detokenize{Important constraints:}\par
\detokenize{- Step indexing is 0-based.}\par
\detokenize{- Select only from the supplied candidate step indices.}\par
\detokenize{- The candidate step indices include every interaction step in the complete trajectory; no steps are filtered out.}\par
\detokenize{- Ground every statement in the task, observations, admissible actions, and complete trajectory.}\par
\detokenize{- Keep episode_skill concise, actionable, and useful at every decision point.}\par
\detokenize{- Express reusable decision rules rather than leaking literal product IDs or other future-only details that were not available at the decision point.}\par
\detokenize{- Use at most one step_skills entry. A failed episode may use an empty object when no unrecovered decisive error can be identified confidently.}\par
\detokenize{- Return only: episode_summary, episode_skill, step_skills.}\par

\detokenize{Return format:}\par
\{\par
\hspace*{1em}\detokenize{"episode_summary": "concise factual summary",}\par
\hspace*{1em}\detokenize{"episode_skill": "global positive workflow",}\par
\hspace*{1em}\detokenize{"step_skills": {{STEP_SKILLS_EXAMPLE}}}\par
\}\par

\detokenize{Episode context:}\par
\detokenize{- Task description: {{TASK_DESCRIPTION}}}\par
\detokenize{- episode_success: {{EPISODE_SUCCESS}}}\par
\detokenize{- Candidate step indices: {{CANDIDATE_STEP_INDICES}}}\par
\detokenize{- Interaction trajectory: {{INTERACTION_TRAJECTORY}}}
\end{minipage}%
}
\endgroup
\caption{WebShop writer prompt template.}
\label{fig:writer-prompt}
\end{figure*}

\section{Implementation and Evaluation Details}
\label{sec:implementation-details}

\subsection{Benchmarks and Result Sources}

\paragraph{ALFWorld and WebShop.}
ALFWorld~\citep{shridhar2020alfworld} evaluates multi-step household tasks through textual observations and actions. We use 2,400 training examples from its training split and evaluate success on all 134 Unseen tasks. WebShop~\citep{yao2022webshop} requires agents to search, filter, configure, and purchase products under compositional constraints. We use 2,400 training examples and all 500 evaluation tasks, reporting normalized task score and exact success rate.

\paragraph{Search QA.}
We follow the Search-R1 training and evaluation protocol~\citep{jin2025search}, using 19,200 training examples and evaluation on Natural Questions, TriviaQA, PopQA, HotpotQA, 2WikiMultiHopQA, MuSiQue, and Bamboogle. We report each dataset's accuracy and their unweighted macro average.

\paragraph{Comparison methods and sources.}
Table~\ref{tab:main} compares ETRO with GRPO, GRPO+OPSD, Skill-SD, RLSD, SDAR, OPID, SEED, and ICSD. All ETRO rows are from our local experiments; every non-ETRO row is an externally reported reference result drawn from the ICSD, OPID, and SEED reports~\citep{lan2026trust,yang2026opid,wu2026seed}, including baseline methods reported within those studies. The local training-dynamics comparison (Figure~\ref{fig:training-dynamics}) and the polarity and truncation ablations (Tables~\ref{tab:polarity-ablation} and~\ref{tab:truncation-ablation}) use Qwen2.5-1.5B-Instruct.

\subsection{Training Configuration}

\paragraph{Actor and writer.}
We train Qwen2.5-Instruct actors at 1.5B, 3B, and 7B scales, using a frozen Qwen3.5-27B writer for all scales. After each completed episode, the writer generates hindsight skills, and the frozen behavior policy re-scores the original responses under the applicable skill contexts using teacher forcing. These operations add training-time computation; skills and writer calls are absent during evaluation.

\paragraph{Rollout batches and sequence limits.}
Each training batch contains 16 tasks for ALFWorld and WebShop or 128 tasks for Search QA, with eight rollouts per task. Thus, an ALFWorld or WebShop batch contains 128 training trajectories. Maximum actor prompt lengths are 2,048 tokens for ALFWorld and 4,096 for WebShop and Search QA, with at most 512 response tokens per turn. Both training and evaluation allow up to 15 interaction turns for ALFWorld and WebShop and four for Search QA. Overlong WebShop prompts are rejected, while Search QA uses left truncation.

\paragraph{Writer decoding.}
The writer uses temperature 0.6, top-$p=0.95$, and top-$k=20$, with a maximum output length of 4,096 tokens and a maximum context length of 65,536 tokens. Thinking is disabled.

\paragraph{Optimization and actor decoding.}
The actor uses AdamW with learning rate $10^{-6}$, betas $(0.9,0.999)$, and weight decay 0.01. We use one PPO epoch per collected batch, a clipping parameter of 0.2, and a KL-loss coefficient of 0.01. The WebShop PPO minibatch size is 64, with a per-GPU microbatch size of eight. Actor rollout decoding uses temperature 1.0 and top-$p=1$.

\paragraph{OPD weighting and loss aggregation.}
ETRO adds the routed OPD token advantages to the GRPO advantages with a shared coefficient $\lambda=0.001$ for both skill polarities, rather than introducing a separate cross-entropy loss. We use \texttt{token-mean} aggregation over all valid actor response tokens, excluding environment observations and padding. Tokens without OPD supervision remain in the denominator and retain their GRPO advantages. The suffix-weight comparison uses this same denominator for every $\alpha$, without separately normalizing by OPD-active or supervised suffix tokens.

\subsection{Evaluation Protocol}

\paragraph{WebShop decoding and checkpoint selection.}
WebShop validation uses sampled decoding at temperature 0.4. The local 1.5B training-dynamics comparison runs for 160 optimization steps and reports each method's final checkpoint rather than its best intermediate result; ETRO reaches 80.8\% success at step 160.

\paragraph{Search QA evaluation.}
We use \texttt{intfloat/e5-base-v2} and a FAISS flat index over \texttt{wiki-18}, retrieving three passages per query. Exact-match evaluation extracts the last \texttt{<answer>} span, lowercases it, removes punctuation and English articles, and collapses whitespace before comparison with accepted answers.

\section{Additional Analysis Experiments}
\label{sec:additional-analysis}

\subsection{Regional Alignment Diagnostic}
\label{sec:alignment-protocol}

\paragraph{Rollout collection and sample composition.}
For each task prompt $x$, the frozen GRPO-trained actor samples eight complete trajectories without hindsight skills. These are eight rollouts from the task's initial state, not eight continuations from each recorded turn. Both successful and failed trajectories contribute to return estimation. Only turns from failed trajectories with a valid writer-estimated first unrecovered decisive error $k_i$ enter the regional correlations in Figure~\ref{fig:preliminary-alignment}; successful turns are not relabeled as pre-error turns. The 100-trajectory writer audit in Section~\ref{sec:analysis} is a separate analysis, not the stated sample size of this diagnostic.

\paragraph{Outcome returns and action pooling.}
The terminal outcome is $R_i\in\{0,1\}$, with success scored as one and failure as zero. With no intermediate outcome reward, an occurrence's discounted future return is $G_{i,t}=\gamma^{T_i-t}R_i$, where $\gamma\in[0,1]$ is the discount factor. Let $\mathcal{H}_x(o)$ contain all turn occurrences $(i,t)$ among the eight rollouts whose pre-action anchor observation is exactly $o$. Its subset $\mathcal{H}_x(o,a)$ contains occurrences executing the same action $a$. Grouping is within a prompt, uses the executable action rather than the full reasoning response, and pools occurrences regardless of their trajectories' final outcomes or error regions. We estimate an action's return by
\begin{equation}
\widehat Q_x(o,a)=\frac{\sum_{(i,t)\in\mathcal{H}_x(o,a)}G_{i,t}}{|\mathcal{H}_x(o,a)|}.
\label{eq:alignment-action-return}
\end{equation}
The anchor baseline is the occurrence-weighted mean
\begin{equation}
\overline G_x(o)=\frac{\sum_{(i,t)\in\mathcal{H}_x(o)}G_{i,t}}{|\mathcal{H}_x(o)|}.
\label{eq:alignment-anchor-baseline}
\end{equation}
Each retained occurrence receives the action-pooled reference advantage by subtracting the shared-anchor baseline from the pooled action return, without standard-deviation normalization:
\begin{equation}
A^{\mathrm{ref}}_{i,t}=\widehat Q_x(o_{i,t},a_{i,t})-\overline G_x(o_{i,t}).
\label{eq:alignment-reference-advantage}
\end{equation}
This GiGPO-style diagnostic reference is used only for the alignment analysis; \methodname{} is trained with the advantage in Equation~\ref{eq:totaladv}.

\paragraph{Filtering and OPD comparison.}
We discard singleton anchor groups and groups without distinct executable actions. The remaining occurrences contribute to the pooled reference; the plotted subset additionally requires a failed trajectory with a valid $k_i$. For each such turn, we average Equation~\ref{eq:opd-gap} over its reasoning and action tokens separately for each skill polarity. We compute turn-level Spearman correlations for $t<k_i$ (prefix), $t=k_i$ (diagnosed error), and $t>k_i$ (suffix). The plot reports these token-mean gaps before the training coefficient $\lambda$ is applied. Correlations are descriptive: multiple turns can share a prompt or anchor, and Figure~\ref{fig:preliminary-alignment} does not report confidence intervals or post-filter coverage.

\subsection{Controlled Recovery Protocol}
\label{sec:recovery-protocol}

To assess whether suffix truncation improves behavior after a recoverable error, we construct a controlled recovery suite from held-out WebShop tasks. We execute a pre-registered, plausible, recoverable wrong action from a shared reference trajectory, then start each actor from the same perturbed state and collect repeated continuations without hindsight skills. The compared actors retain Positive prefix and Negative error-turn guidance during training but differ in suffix treatment: Suffix Positive continues Positive OPD, Suffix Negative continues Negative OPD, and ETRO retains only GRPO. We measure recovery success, normalized score, the proportion of first continuation actions classified by a blinded evaluator as wrong-branch continuation, and entropy over those first-turn action tokens.

The recovery results are reported in Table~\ref{tab:actor-behavior} and discussed in Section~\ref{sec:analysis}.

\end{document}